\documentclass[sigconf]{acmart}

\renewcommand\footnotetextcopyrightpermission[1]{}

\usepackage{booktabs} 

\usepackage[colorinlistoftodos]{todonotes}
\usepackage{comment}

\graphicspath{{images/}}
\DeclareGraphicsExtensions{.pdf,.jpeg,.png}

\usepackage{multicol}
\usepackage{multirow}

\usepackage{threeparttable}

\usepackage[caption=false,font=footnotesize]{subfig}

\usepackage{xspace}
\usepackage[nolist,nohyperlinks]{acronym}

\acmConference[NIPS2017-DLTP]{Deep Learning: Bridging Theory and Practice}{December 9, 2017}{Long Beach, USA}

\newacro{nln}[\textit{N-light-N}\xspace]{\textit{N-light-N}}

\newacro{scae}[\textsc{SCAE\xspace}]{Stacked Convolutional Auto-Encoder}

\newacro{ae}[\textsc{AE\xspace}]{Auto-Encoder}
\newacroplural{ae}[\textsc{AEs\xspace}]{Auto-Encoders}

\newacro{cae}[\textsc{CAE\xspace}]{Convolutional Auto-Encoder}

\newacro{nn}[\textsc{NN\xspace}]{Neural Network}
\newacroplural{nn}[\textsc{NNs\xspace}]{Neural Networks}

\newacro{aenn}[\textsc{AENN\xspace}]{Auto-Encoder Neural Network}
\newacro{ann}[\textsc{ANN\xspace}]{Artificial Neural Network}

\newacro{cnn}[\textsc{CNN\xspace}]{Convolutional Neural Network}
\newacro{ffcnn}[\textsc{FFCNN\xspace}]{Feed Forward Convolutional Neural Network}

\newacro{dnn}[\textsc{DNN\xspace}]{Deep Neural Network}
\newacroplural{dnn}[\textsc{DNNs\xspace}]{Deep Neural Networks}

\newacro{aec}[\textsc{AEC}]{Auto-Encoder Classifier}
\newacro{ffcnn}[\textsc{FFCNN\xspace}]{Feed Forward Convolution Neural Network}

\newacro{svd}[\textsc{SVD\xspace}]{Singular Value Decomposition}
\newacro{pca}[\textsc{PCA\xspace}]{Principal Component Analysis}
\newacro{apca}[\textsc{APCA\xspace}]{Activated Principal Component Analysis}
\newacro{cpca}[\textsc{CPCA\xspace}]{Converted Principal Component Analysis}
\newacro{lda}[\textsc{LDA\xspace}]{Linear Discriminant Analysis}
\newacro{relu}[\textsc{ReLU\xspace}]{Rectified Linear Units}

\begin{document}
\title[A Pitfall of Unsupervised Pre-Training]{A Pitfall of Unsupervised Pre-Training}

\subtitle{Extended Abstract}

\author{Michele Alberti, Mathias Seuret, Rolf Ingold,  Marcus Liwicki}

\affiliation{%
    \department{Document Image and Voice Analysis Group (DIVA)}
    \institution{University of Fribourg}
    \streetaddress{Bd. de Perolles 90}
    \city{Fribourg} 
    \state{Switzerland} 
}
\email{{firstname}.{lastname}@unifr.ch}

\renewcommand{\shortauthors}{M. Alberti et al.}

\begin{abstract}

The point of this paper is to question typical assumptions in deep learning and suggest alternatives.
A particular contribution is to prove that even if a \ac{scae} is good at reconstructing pictures, it is not necessarily good at discriminating their classes. %
When using \aclp{ae}, intuitively one  assumes that features which are good for reconstruction will also lead to high classification accuracy. %
Indeed, it became research practice and is a suggested strategy by introductory books. %
However, we prove that this is not always the case. %
We thoroughly investigate the quality of features produced by \acp{scae} when trained to reconstruct their input. %
In particular, we analyze the relation between the reconstruction and classification capabilities of the network, if we were to use the same features for both tasks. %
Experimental results suggest that in fact, there is no correlation between the reconstruction score and the quality of features for a classification task. %
This means, more formally, that the sub-dimension representation space learned from the \ac{scae} (while being trained for input reconstruction) is not necessarily better separable than the initial input space. 
Furthermore, we show that the reconstruction error is not a good metric to assess the quality of features, because it is biased by the decoder quality. %
We do not question the usefulness of pre-training, but we conclude that aiming for the lowest reconstruction error is not necessarily a good idea if afterwards one performs a classification task.

\end{abstract}

\maketitle

\section{Introduction}

\begin{figure}[!b]
  \begin{center}
    \includegraphics[width=\columnwidth]{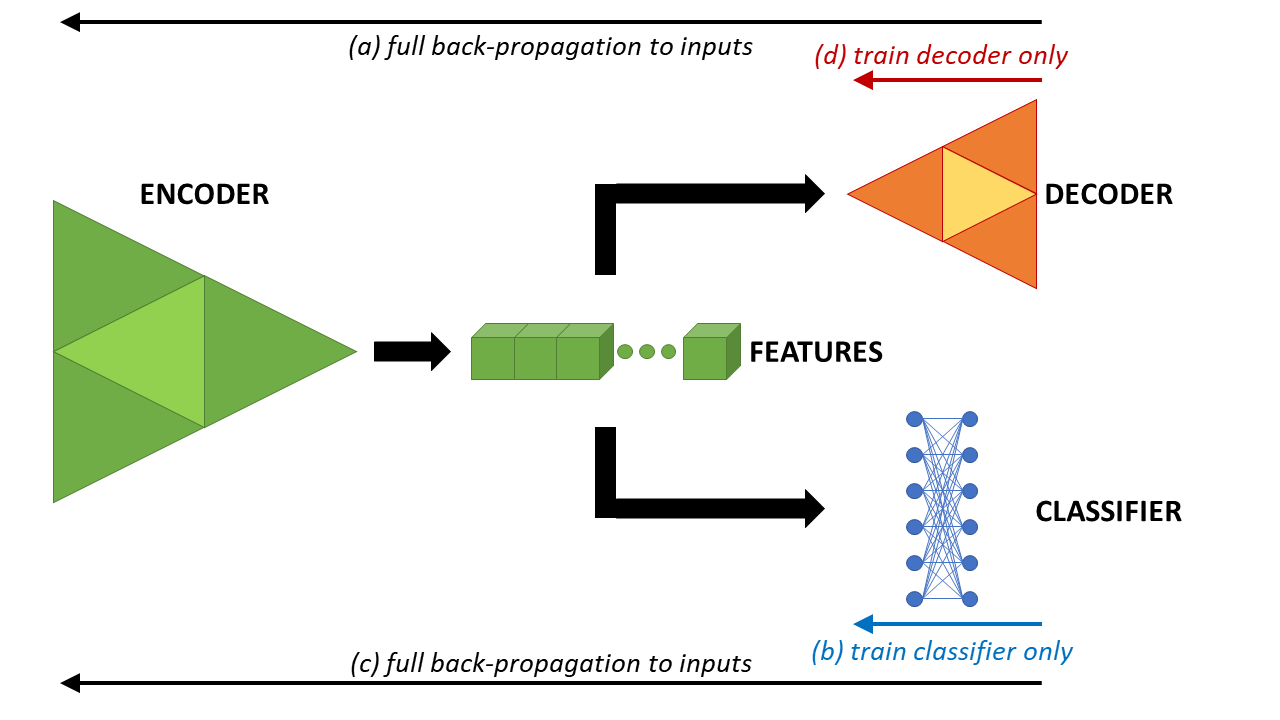}
    \caption{Visual representation of the depth of back propagation when: training the \ac{scae} to reconstruct its input (a), training only the classifier (b), training the classifier with fine-tuning the whole \ac{scae} (c) and training just the decoder part (d).}
    \label{fig:methodology}
  \end{center}
\end{figure}

\begin{figure*}[!t]
\centering
\begin{tabular}{ccc|ccc|ccc}
\includegraphics[width=0.085\textwidth]{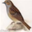} &
\includegraphics[width=0.085\textwidth]{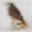} &
\includegraphics[width=0.085\textwidth]{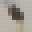} &
\includegraphics[width=0.085\textwidth]{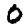} &
\includegraphics[width=0.085\textwidth]{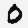} &
\includegraphics[width=0.085\textwidth]{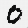} &
\includegraphics[width=0.085\textwidth]{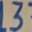} &
\includegraphics[width=0.085\textwidth]{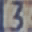} &
\includegraphics[width=0.085\textwidth]{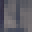} \\

\includegraphics[width=0.085\textwidth]{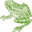} &
\includegraphics[width=0.085\textwidth]{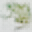} &
\includegraphics[width=0.085\textwidth]{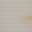} &
\includegraphics[width=0.085\textwidth]{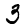} &
\includegraphics[width=0.085\textwidth]{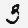} &
\includegraphics[width=0.085\textwidth]{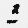} &
\includegraphics[width=0.085\textwidth]{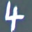} &
\includegraphics[width=0.085\textwidth]{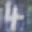} &
\includegraphics[width=0.085\textwidth]{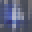} \\

\includegraphics[width=0.085\textwidth]{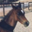} &
\includegraphics[width=0.085\textwidth]{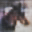} &
\includegraphics[width=0.085\textwidth]{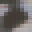} &
\includegraphics[width=0.085\textwidth]{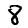} &
\includegraphics[width=0.085\textwidth]{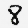} &
\includegraphics[width=0.085\textwidth]{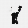} &
\includegraphics[width=0.085\textwidth]{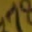} &
\includegraphics[width=0.085\textwidth]{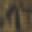} &
\includegraphics[width=0.085\textwidth]{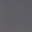} \\

Original & Before & After & Original & Before & After & Original & Before & After \\
\multicolumn{3}{c}{\textbf{CIFAR}} & \multicolumn{3}{c}{\textbf{MNIST}} & \multicolumn{3}{c}{\textbf{SVHN}} \\

\end{tabular}
\caption{Visualization of \ac{scae} reconstructions before and after fine-tuning for the datasets of object recognition (CIFAR) and digit recognition (MNIST and SVHN). For each class in the three datasets is shown a representative sample chosen randomly. Notice how the reconstruction capabilities of the networks are significantly lower after fine tuning for classification for both CIFAR and SVHN but only in a much smaller magnitude for MNIST.}
\label{fig:results_recoded_imageBased}
\end{figure*}

\begin{figure*}[!t]
    \centering
    \subfloat[Original image. ]{\includegraphics[width=.5\columnwidth]{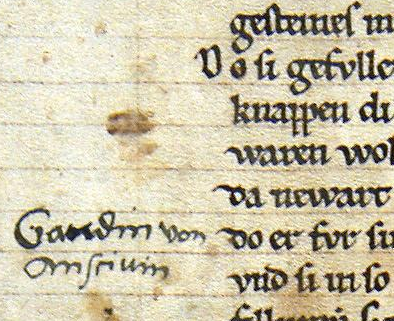}\label{sfig:results_d-007}}
    \hfil
    \subfloat[Initial SCAE reconstruction.  Average $L_2$ error: $0.31$]{\includegraphics[width=.5\columnwidth]{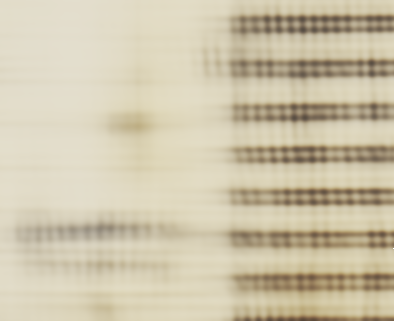}\label{sfig:results_SCAE-recoded}}
    \hfil
    \subfloat[Reconstruction after fine tuning the SCAE.  Average $L_2$ error: $0.55$]{\includegraphics[width=.5\columnwidth]{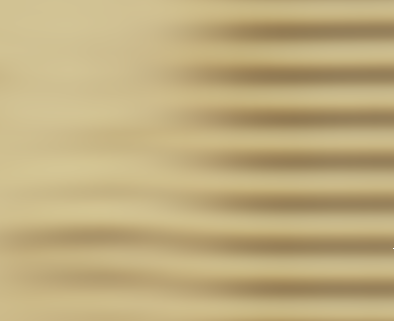}\label{sfig:results_FT-recoded}}
    \hfil
    \subfloat[Reconstruction after training the decoder only.  Average $L_2$ error: $0.45$]{\includegraphics[width=.5\columnwidth]{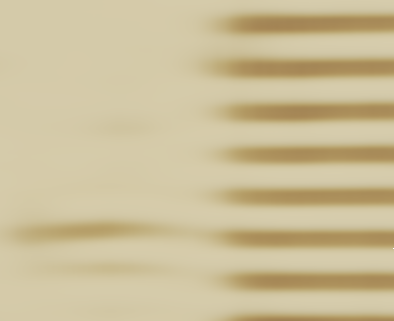}\label{sfig:results_TDO-recoded}}
      
    \caption{
        A patch of an image of the Parzival dataset at the different stages of the experiments pipeline. %
        After fine tuning the classifier the classification accuracy is higher but so is the reconstruction error. %
        Notice how in (d) the stains from original image are no longer reconstructed in contrast with (b) where they are still present. %
    }
    \label{fig:results_recoded}
\end{figure*}

In recent years the knowledge on \ac{dnn} made huge steps forward, yet there is still no clear and exhaustive understanding of when and why a deep model works. %
This makes it difficult to discriminate reliable good practices from techniques that might work, but it is not clear under which circumstances they perform well. %
\par %
Too often authors are applying a particular technique just because ``once it worked'' or because modern Deep Learning frameworks allow to use many tools ``out of the box'' removing the requirement of understanding how they work.  %
This, unfortunately, often leads to commonly agreed practices without proper scientific verification. %

\subsection*{Importance of this Work}

People who are proficient in machine learning know that --- despite there have been many very successful applications --- features that are good/bad for reconstruction are not necessarily good/bad for classification, as these tasks are different.
However, to the best of our knowledge, there is still a lack of literature on the subject.
Moreover, people who are beginners/not much used to machine learning might probably focus much on the reconstruction accuracy, which could be a waste of time.
In fact, not only many prominent online lectures \cite{rogergrosse2017,colinmcdonnell2016,reimers2015} but also recently published books \cite{goodfellow2016} still suggest to use the compressed representation of \acp{ae} to enable better performance on classification tasks.

\subsection*{Main Contribution}

This investigation does not have the purpose of providing technical novelty but instead rejecting a commonly agreed assumption on a non-novel and widely used approach. %
Thus, the contribution of this paper is not a novel method but a better understanding of the foundations of many existing methods. %
Specifically, we reject the thought that using auto-encoder sub-dimensional features --- learned by reconstructing the input --- for classification tasks, is always good idea. %

\subsection*{Related work}

With their recent success in Computer Vision, the research area of deep learning becomes more and more popular. %
A widely known technique for fast learning of very deep networks is layer-wise training. %
This technique has been introduced by Ballard \cite{ballard1987modular} in 1987 and became popular after its successful application in the last decade by many different authors. %
Among them, there are Hinton et al.~\cite{hinton2006} who used it to train deep belief networks in 2006, Bengio et al.~\cite{bengio2007greedy} who in 2007 showed how deep architectures are more efficient than shallow ones for difficult problems and Lee et al.~\cite{lee2009convolutional} who introduced convolution and probabilistic max-pooling in deep belief networks in 2009. %
These innovations contributed to the development of a new architecture paradigm which is obtained by literally stacking and convolving \acp{ae}. %
The concept of stacking \acp{ae} became popular with Vincent et al.\ \cite{vincent2010stacked} in 2010. %
One year later Masci et al.\ \cite{masci2011} put convolution into play and introduced the \ac{scae} architecture. %
In the more recent years several authors made used of this paradigm in very different contexts. %
For example Chen et al. \cite{chen2015page} applied it in historical documents image layout analysis, Tan et al.\ \cite{tan2014stacked} used it for steganalysis of digital images and Leng et al.\ \cite{Leng2015} for 3D object retrieval.
All these work share the common practice to use \acp{ae} as feature extractors for performing classification tasks.
However, by doing this, it is inherently assumed that a good \ac{ae} will lead to a good classification.
Here we continue the work of Wei et al.~\cite{wei} on analyzing the quality of automatically/deeply learned features by questioning the correlation between their reconstruction and classification abilities.

\section{Reconstruction Error}
\label{sct:rerror}

In this section we analyze the reliability of reconstruction error used as metric to determine the quality of features produced by \acp{scae}. %
A common way to evaluate the quality of an \ac{ae} (or to evaluate its learning status) is to look at its reconstruction error. %
Recall that the purpose of an \ac{ae} is being able to reconstruct an input $\vec{x}$ from its encoded representation $\vec{y}$ as in Equation~\ref{eq:ae_idea}:

\begin{equation} \label{eq:ae_idea}
\vec{y}=E\left(\vec{x}\right)~,~\vec{x}^\prime=D\left(\vec{y}\right)~|~\vec{x}\approx \vec{x}^\prime \quad \text{where}\quad |\vec{y}| \leq |\vec{x}|
\end{equation}

where $E$ and $D$ are the encoding/decoding functions. %
The reconstruction error $s$ is computed by measuring the distance between the input and the reconstructed output with a distance function $L$, as shown in Equation~\ref{eq:reconstruction_error}: 

\begin{equation} \label{eq:reconstruction_error}
s = L(\vec{x}^\prime,\vec{x})
\end{equation}

Using the reconstruction error to evaluate the classification capabilities of the features of an \ac{ae} or a \ac{scae} is unsafe for three main reasons: \\

\begin{enumerate}
    \item There is no mathematical background supporting the hypothesis that good features for input reconstruction are inherently good for classification purposes. %
    \acp{ae} are trained to reconstruct their input and could learn to ignore aspects of the input which could be critical for a successful classification task. \\ 
    
    \item The \ac{ae} is not necessarily learning a meaningful representation of the input. In fact, an extreme example would be when the identity function\footnote{Under the assumption $|\vec{x}| \leq |\vec{y}|$ this is not only possible but also very likely to happen.} is learned from the \ac{ae}. In this case, the reconstruction error is $0$ but the quality of the features vector $\vec{y}$ is poor as there is no advantage from feeding them rather than the raw input to a classifier. \\
    
    \item After substituting~\ref{eq:ae_idea} in~\ref{eq:reconstruction_error} we obtain that the reconstruction error is computed as: 
    
    \begin{equation}\label{eq:reconstrcution_error}
    s = L(D(E(\vec{x})),\vec{x})
    \end{equation}
    
    That is, the decoding function $D(\vec{y})$ is taken into account and affects the reconstruction error $s$. This means that a bad decoding function $D$ can shadow an high quality features vector $\vec{y}$. Furthermore, when using the \ac{scae} as features extractor, the decoding function is not even used as the encoded array $\vec{y}$ (and not $\vec{x}^\prime$) is forwarded to following layers.  
\end{enumerate} 

In conclusion, both from the mathematical and intuitive points of view, this metric is potentially not giving an insight on the quality of the features vector $\vec{y}$ and its reliability is definitely jeopardized by the quality of the decoding function $D(\vec{y})$. In Section~\ref{sct:results} we show that this analysis find correspondence in practice.

\section{Experiments Setting}
\label{sct:experiments_setting}

\begin{table*}[!t]
\centering
\begin{tabular}{@{}llrrrrrlrrrrr@{}}
\toprule
\multicolumn{2}{l}{} & \multicolumn{5}{c}{Before fine-tuning} & & \multicolumn{5}{c}{After fine-tuning} \\ \cmidrule{3-7} \cmidrule{9-13}
\multicolumn{2}{l}{} & \textbf{EUC} & \textbf{SOI} & \textbf{NOC} & \textbf{E94} & \textbf{MA} &       & \textbf{EUC} & \textbf{SOI} & \textbf{NOC} & \textbf{E94} & \textbf{MA}\\ \midrule

\multirow{2}{*}{MNIST}          & NN & $-0.75^*$& $-0.75^*$	& $-0.75^*$	& $-0.75^*$	& NaN$\hphantom{*}$   &       & $-0.93\hphantom{*}$	& $-0.74^*$	& $-0.88\hphantom{*}$	& $-0.93\hphantom{*}$	& $-0.92\hphantom{*}$\\  
                                & LL & $0.67^*$	& $0.67^*$	& $0.68^*$	& $0.67^*$	& NaN$\hphantom{*}$   &       & $-0.76^*$	& $0.68^*$	& $-0.83^*$	& $-0.76^*$	& $-0.93\hphantom{*}$\\  
\multirow{2}{*}{SVHN}           & NN & $-0.54^*$& $-0.65^*$	& $-0.55^*$	& $-0.61^*$	& $-0.72\hphantom{*}$   &       & $-0.71^*$	& $-0.68^*$	& $-0.66^*$	& $-0.71^*$	& $-0.21^*$\\ 
                                & LL & $0.43^*$	& $0.63^*$	& $0.55^*$	& $0.54^*$	& $0.73\hphantom{*}$    &       & $0.20^*$	& $0.42^*$	& $0.25^*$	& $0.34^*$	& $0.63^*$\\  

\multirow{2}{*}{CIFAR}          & NN & $-0.78\hphantom{*}$	& $-0.80\hphantom{*}$	& $-0.80$	& $-0.82\hphantom{*}$	& $-0.87\hphantom{*}$   &       & $-0.95\hphantom{*}$	& $-0.64^*$	& $-0.72\hphantom{*}$	& $-0.95\hphantom{*}$	& $-0.16^*$\\  
                                & LL & $0.40^*$	& $0.47^*$	& $0.49^*$	& $0.48^*$	& $0.56^*$  &       & $0.16^*$	& $0.31^*$	& $0.11^*$	& $0.20^*$	& $0.44^*$\\  

\multirow{2}{*}{PARZIVAL}       & NN & $-0.47\hphantom{*}$	& $-0.36\hphantom{*}$	& $-0.44\hphantom{*}$	& $-0.48\hphantom{*}$	& $-0.51\hphantom{*}$   &       & $0.42\hphantom{*}$	& $0.45\hphantom{*}$	& $0.44\hphantom{*}$	& $0.41\hphantom{*}$	& $0.43\hphantom{*}$\\  
                                & LL & $-0.08^*$& $0.00^*$	& $0.04^*$	& $-0.04^*$	& $-0.05^*$ &       & $0.68\hphantom{*}$	& $0.81\hphantom{*}$	& $0.79\hphantom{*}$	& $0.75\hphantom{*}$	& $0.82\hphantom{*}$\\                                 
\multirow{2}{*}{SAIN GALL }     & NN & $-0.56\hphantom{*}$	& $-0.60\hphantom{*}$	& $-0.59\hphantom{*}$	& $-0.56\hphantom{*}$	& $-0.57\hphantom{*}$   &       & $-0.85\hphantom{*}$	& $-0.88\hphantom{*}$	& $-0.87\hphantom{*}$	& $-0.86\hphantom{*}$	& $-0.68\hphantom{*}$\\ 
                                & LL & $0.13\hphantom{*}$	& $0.39\hphantom{*}$	& $0.44\hphantom{*}$	& $0.14\hphantom{*}$	& $0.29\hphantom{*}$    &       & $0.38\hphantom{*}$	& $0.65\hphantom{*}$	& $0.73\hphantom{*}$	& $0.43\hphantom{*}$	& $0.25\hphantom{*}$\\                      
\multirow{2}{*}{GW }            & NN & $-0.38\hphantom{*}$	& $-0.35^*$	& $0.27^*$	& $-0.38\hphantom{*}$	& NaN$\hphantom{*}$   &       & $0.44\hphantom{*}$	& $0.44\hphantom{*}$	& $0.31^*$	& $0.44\hphantom{*}$	& $-0.21^*$\\  
                                & LL & $0.02^*$	& $-0.03^*$	& $0.12^*$	& $0.02^*$	& NaN$\hphantom{*}$   &       & $-0.60\hphantom{*}$	& $-0.63\hphantom{*}$	& $-0.70\hphantom{*}$	& $-0.60\hphantom{*}$	& $-0.31^*$\\                           
\bottomrule
\end{tabular}
    \caption{
        Correlation between the reconstruction error and the accuracy obtained on the test for each dataset after training the classifier normally. %
        The star exponent ($^*$) denotes a non significant result (p-value above $0.05$). %
        NN and LL indicate the type of the layers of the classifier: Neural Layer and Linear Layers respectively. %
        The NaN values for the Mahalanobis distance on MNIST and GW datasets are caused by a singular correlation matrix between the input and the reconstructed patch. %
        This is most likely happening because of the many white pixels found in these two datasets. %
    }
\label{tab:results_correlation}
\end{table*}

Our experiments are composed of four phases (see Figure~\ref{fig:methodology}): %

\begin{itemize}
    \item[(a)] Train the \acp{scae} to reconstruct its input by minimizing the reconstruction error. %
    \item[(b)] Use \ac{scae} extracted features to train a classifier (see \ref{ssct:task_dataset} for details on which tasks). %
    \item[(c)] Again, use \ac{scae} extracted features to train a classifier, but this time with fine-tuning the whole \acp{scae} (all layers). %
    \item[(d)] Train the \acp{scae}'s decoders only\footnote{This is necessary to enable the comparison of the reconstruction error before and after fine tuning.}. %
\end{itemize}

Step (a) is the known unsupervised layer-wise pre-training process. %
In step (b) we train a classifier on the features extracted from the \ac{scae} to measure their classification capabilities. %
In step (c) allowing the feature extractor to ``fit'' to the classifier instead of keeping it statically defined sounds like a reasonable idea\footnote{This is not novel and its a well known practice in the field.}: but what does this mean from the features point of view? %
If a fully trained (hence converged) \ac{scae} can be still modified such to increase the performance of a classifier, we can safely derive that the former features were not optimal for that task. %
As a matter of fact, the \ac{scae} has been taught to reconstruct the input, not to extract the best features for classification. %
Conversely, we want to investigate if the additional fine-tuning for the classification task harms the reconstruction abilities of the learned representations. %
Therefore with step (d) we compare the reconstruction capabilities before and after fine-tuning for classification.

\subsection{Classification Tasks and Datasets}
\label{ssct:task_dataset}

In the context of this work, we would need a single negative example to disprove the generality of the assumption that features good for reconstruction are inherently good for classification. %
We however run a thorough investigation on different tasks belonging to different domains, such as digit recognition (MNIST \cite{lecun1998gradient} and SVHN~\cite{netzer2011reading}), object recognition (CIFAR10 ~\cite{krizhevsky2009learning}) and historical documents image segmentation at pixel level (IAM-HistDB: Parzival, Sain Gall and G. Washington~\cite{fischer2011,fischer2012}).

\subsection{Architecture}

We use the standard version of the \ac{scae} architecture~\cite{seuretalberti2016nln} --- without modifying it --- because we want to make a point on a general setting and not on a specific case. %
In fact, to increase generality we did not stick to one architecture but we adapted multiple configurations similar to those used in recent work~\cite{seuretalberti2017pca,alberti2017lda} which have up to 6 layers and at most $\sim$4K parameters. %
In total we measured the results on 10 different \acp{scae} configurations (different hyperparameters, architecture, initialization technique, ... ) and tested each of them once with a linear classifier and once with a neural classifier made of 3 fully connected layers with a few hundreds of neurons each. %
As we are measuring the classification performances only once the network is converged (applies both to classifiers and \ac{scae}), the exact values of some hyper-parameters (e.g the learning rate) are not relevant and are therefore omitted. %

\section{Results Discussion}
\label{sct:results}

\begin{figure}[t]
    \centering
    \includegraphics[width=\linewidth]{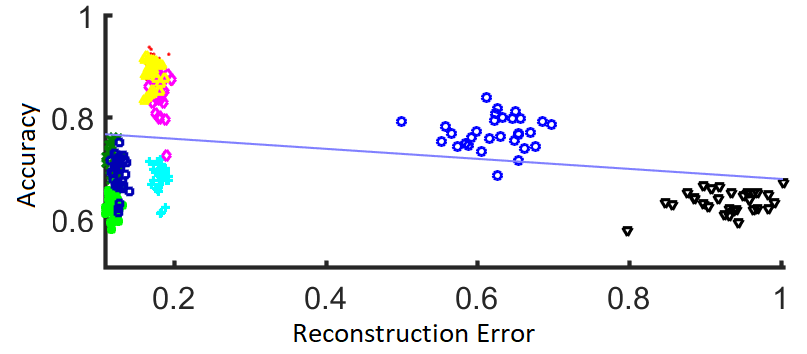}
    \caption{
        In the above plot there are the performance of different configurations on the Sain Gall dataset. %
        Different color correspond to a different configuration of the \ac{scae}. %
        Each point represent the result of the \ac{scae}-classifier pair on a specific page belonging to the test set. %
        All the pairs shown here have linear layers in the classifier. %
        The reconstruction error is computed with the Euclidean distance metric. %
        Notice how the highest accuracy is not achieved with the smallest reconstruction error (yellow/pink points). %
        The blue line is the visualization of the correlation. 
        }
    \label{fig:[STG]LL-ALL-EUC}
    \vspace{-0.4cm}
\end{figure}

After training the \ac{scae} and using its features to train the classifier (phases \textit{a} and \textit{b}, see Section~\ref{sct:experiments_setting}) we can measure the relationship between the reconstruction error and the classification accuracy. %
In Figure~\ref{fig:[STG]LL-ALL-EUC}, one can see that the highest classification accuracy has not been obtained by the \ac{scae} which had the lower reconstruction error. %
In fact, some of the \acp{scae} that had a very low reconstruction error produced features with which the classifier reached a very low accuracy (e.g light green points in Figure~\ref{fig:[STG]LL-ALL-EUC}).
This already proves that aiming for the lowest reconstruction error is not necessarily a good idea if afterwards one performs a classification task. %

\par %
We measured the correlation coefficients between the reconstruction error and the classification accuracy for all configurations on all different datasets, then, fine-tuned the classifier and the decoder parts (phases \textit{c} and \textit{d}, see Section~\ref{sct:experiments_setting}) and finally measured the correlation once again. %
We used different distance metrics to show that this is not influencing the results. %
Specifically: Euclidean distance (EUC), Scaled Offset Invariant (SOI), Normalized Correlation (NOC), $\Delta$ E94 and Mahalanobis. %
The results are summarized in Table~\ref{tab:results_correlation}. %
As expected, the absence of a pattern or a general trend denotes that there is no correlation between these two values. %
For instance, notice how only some datasets present different trend before/after fine-tuning, e.g. the George Washington dataset has flipped signs between the two tables. %
There is no coherent trend for the correlation, even in the same domain. %
For example, observe how the three image segmentation datasets have completely different numbers --- especially after fine-tuning. %

\par %
In Figure~\ref{fig:results_recoded} we visualized the experiments pipeline to make it easier to interpret. %
We show a patch of an image of the Parzival dataset at the different stages of the experiments. %
On this patch, fine-tuning the classifier/decoder increases both the classification accuracy and the reconstruction error. %
In fact, the quality of the final reconstruction (\ref{sfig:results_TDO-recoded}) is worse than the initial \ac{scae} reconstruction (\ref{sfig:results_SCAE-recoded}) and, as expected, before training the decoder (\ref{sfig:results_FT-recoded}) is even worse. %
This is also observable in their Euclidean reconstruction error which is reported for each stage. 

\par %
Additionally, we noticed how the stains in the original image (\ref{sfig:results_d-007}) are not reconstructed by the fine-tuned network (\ref{sfig:results_TDO-recoded}). %
This is not surprising because in historical document image analysis the stains are typically considered source of noise and do not contain useful information to discriminate a pixel class. %
In a way, the fine-tuned network applies a ``denoising'' effect, w.r.t the information useful to discriminate classes. %

\par %
Another observation we made is that the fine-tuning process can lead to features which are not linked to visual appearance as the decoder is unable to reconstruct the SCAE input after fine-tuning. %
This is visible in Figure~\ref{fig:results_recoded_imageBased}: on the frog image for CIFAR, the number $7$ for the SVHN and with a minor magnitude on the numbers $3$ and $8$ for MNIST. %

\section{Conclusion and Outlook}

In this paper we questioned a typical assumptions made in deep learning. %
We reject the Hypothesis that \textit{``Features good for reconstruction are inherently good for classification tasks.''} by showing that thers is no correlation between reconstruction and classification abilities of automatically/deeply learned features of \acp{scae}. 
\par %
This work contributes towards advancing knowledge not only on the \ac{scae} paradigm, but also on the general field of unsupervised feature learning. %
While this idea is getting more and more attention from the researchers in the field of machine learning and deep learning, we strive better understanding the internals, the found representations, and the classification potential. %
\par %
Finally, we conclude that one should not rely on reconstruction error to evaluate the quality of the features produced by a \ac{scae} for a classification task, but rather aim for an alternative and independent investigation of the network classification abilities. %

\begin{acks}
The work presented in this paper has been partially supported by the HisDoc III project funded by the Swiss National Science Foundation with the grant number $205120$\textunderscore$169618$.
\end{acks}

\bibliographystyle{ACM-Reference-Format}
\bibliography{biblio} 

\end{document}